 \documentclass[final,3p,times,authoryear]{elsarticle}

\usepackage{amssymb}
 \usepackage{amsthm}
 \usepackage{amsmath}
 \usepackage{color, colortbl}
 \usepackage{amsmath}
\usepackage{siunitx}
\usepackage{todonotes}
\usepackage{tabularx}

\definecolor{light-gray}{gray}{0.9}

\usepackage{framed} 

\journal{Urban Climate}

\begin{document}

\begin{frontmatter}

\title{Sky pixel detection in outdoor imagery using an adaptive algorithm and machine learning}

\author[melb,monash,crc]{Kerry~A.~Nice\corref{cor1}}
\ead{kerry.nice@unimelb.edu.au}
\author[melb]{Jasper S. Wijnands}
\author[asu]{Ariane Middel}
\author[cis]{Jingcheng Wang}
\author[cis]{Yiming Qiu}
\author[cis]{Nan Zhao}
\author[melb]{Jason Thompson}
\author[melb]{Gideon D.P.A. Aschwanden}
\author[melb]{Haifeng Zhao}
\author[melb,eng]{Mark Stevenson}
\cortext[cor1]{Principal corresponding author}
\address[melb]{Transport, Health, and Urban Design Hub, Faculty of Architecture, Building, and Planning, University of Melbourne, Australia.}
\address[cis]{School of Computing and Information Systems, University of Melbourne, Australia.}
\address[eng]{Melbourne School of Engineering; and Melbourne School of Population and Global Health, University of Melbourne, Australia.}
\address[monash]{School of Earth, Atmosphere and Environment, Monash University, Australia.}
\address[crc]{Cooperative Research Centre for Water Sensitive Cities, Melbourne, Australia.}
\address[asu]{School of Arts, Media and Engineering (AME), School of Computing, Informatics, and Decision Systems Engineering (CIDSE), Arizona State University.}

\begin{abstract}

Computer vision techniques enable automated detection of sky pixels in outdoor imagery. In urban climate, sky detection is an important first step in gathering information about urban morphology and sky view factors. However, obtaining accurate results remains challenging and becomes even more complex using imagery captured under a variety of lighting and weather conditions. 

To address this problem, we present a new sky pixel detection system demonstrated to produce accurate results using a wide range of outdoor imagery types. Images are processed using a selection of mean-shift segmentation, K-means clustering, and Sobel filters to mark sky pixels in the scene. The algorithm for a specific image is chosen by a convolutional neural network, trained with 25,000 images from the Skyfinder data set, reaching 82\% accuracy for the top three classes. This selection step allows the sky marking to follow an adaptive process and to use different techniques and parameters to best suit a particular image. An evaluation of fourteen different techniques and parameter sets shows that no single technique can perform with high accuracy across varied Skyfinder and Google Street View data sets. However, by using our adaptive process, large increases in accuracy are observed. The resulting system is shown to perform better than other published techniques.

\end{abstract}


\end{frontmatter}

\section{Introduction}\label{sec:introduction}
Sky pixel detection in images is an ongoing computer vision challenge with a large range of applications such as autonomous vehicle or drone navigation \citep{Shen2013}, real-time weather classification \citep{Roser2008}, image editing \citep{Laffont2014,Tao2009}, sky replacement \citep{Tsai2016}, and scene parsing \citep{Tighe2010,Hoiem2005}. It is also an important tool in urban climate research. The use of fisheye photography to calculate sky view factors (SVF), the fraction of sky visible to a point, at individual locations has long been used in urban climate. Numerous techniques exist to process this sort of imagery \citep{Grimmond2001,Chapman2004,Ali-Toudert2007}.

In computer vision research, sky detection techniques have largely followed two main paths, either finding the pixels associated with the sky on a pixel by pixel basis or by finding a sky/ground boundary and labelling the sky as everything above that boundary. The first approach focuses on finding individual pixels associated with the sky. \cite{Luo2002} employed a physics-based approach using the changes of sky colours from zenith to horizon. \cite{Gallagher2004} generated sky pixel probability maps based on colour values and a two-dimensional polynomial for each colour channel. \cite{Zafarifar2007} added texture, gradients, and vertical position to colour values to generate their probability map, however, only blue sky is detected accurately with clouds marked with low probabilities. \cite{Schmitt2009} added in an analysis of position and shape and were reportedly able to also accurately perform sky detection under cloudy conditions. 

Other approaches have focused on finding a sky/ground boundary. Straight lines were first used to define a horizon located via an energy function \citep{Ettinger2003}. Using an improved energy function optimisation and gradient information from the image, \cite{Shen2013} allowed the horizon line to follow the boundary instead of being restricted to a straight line. Additional variations allowed increasingly difficult sky regions (i.e. regions separated from the main sky region by buildings, flags, or other obstructions) to be detected \citep{Zhijie2014,Zhijie2015}. Another approach, not specifically designed for sky pixel identification, attempts to classify around a dozen classes (such as sky, buildings, trees, and cars) in outdoor imagery through semantic segmentation using trained convolutional neural networks (CNN), such as SegNet \citep{Badrinarayanan2017}, and other variations \citep{Holder2016,Middel2019}.

Comparing existing approaches is difficult, as most studies do not report accuracy metrics, and when they do, they are reported using different metrics and benchmark data sets. \cite{Luo2002} reports 90.4\% correct detection of blue sky pixels and 13\% misclassifications. \cite{Chapman2004} reports a root-mean-squared error (RMSE) of 0.06 in marking SVF in fisheye images. \cite{Schmitt2009} reports an accuracy of 0.90 for 80\% of their validation images and an accuracy of 0.95 for 75\% of the images. \cite{Liang2017}, using SegNet, reports an accuracy of 0.96 for sky pixel identification. \cite{Shen2018}, using an off-the-shelf version of SegNet, reports an accuracy of 0.83 with lateral views. Finally, \cite{Middel2019} reports an accuracy of 0.95 with lateral views.

An effort is ongoing to provide worldwide databases of standardised urban morphology information \citep{Ching2018,Ching2019}. Now with the widespread availability of urban imagery, an opportunity exists to expand the range of research that can be conducted without the requirement of manually collecting urban morphology parameters and to accelerate the population of databases such as the World Urban Database and Portal Tool (WUDAPT) \citep{Mills2015}.

Studies have started to utilise automated methods to build SVF data sets from Google Street View (GSV) imagery \citep{Middel2018,Gong2018}. The results are promising but show some accuracy problems. Urban areas with large numbers of street trees in particular are cited by \cite{Gong2018} as a key source of inaccuracies in their system. In addition, these systems are highly dependent on GSV imagery, which as of 2018 has become more restrictive to license and expensive to obtain. This necessitates the need to expand the type of imagery used, imagery that unlike GSV, might vary more in lighting, weather conditions, camera angles, and aspect ratios. 

With these factors in mind, we present a sky pixel detection system that has been tested using various types of outdoor imagery collected under a wide range of lighting and weather conditions, camera angles, and aspect ratios. This system, built on artificial intelligence, is adaptive and uses a range of algorithms and combinations of parameters to locate the sky pixels to ensure the highest accuracy for each individual class of images. We evaluate a number of existing and new techniques for sky pixel classification and demonstrate that this new adaptive system performs with greater accuracy than any of these individual techniques on their own.

\section{Methods}\label{sec:Methods}
\begin{figure}
\centering    
\fbox{
\includegraphics[trim={0 0 0 0},clip,scale=0.20]{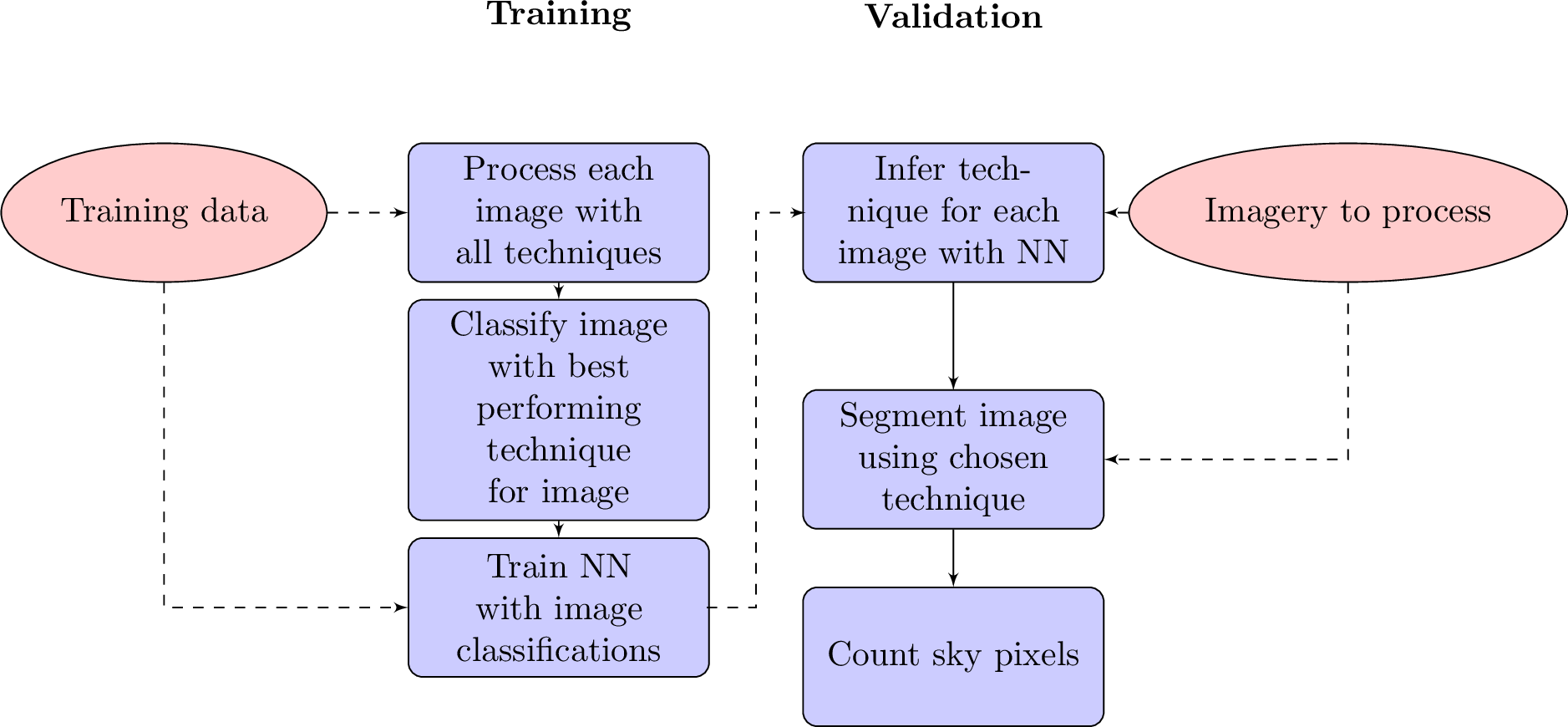}
}
\caption{\bf Process flow of training and validation steps.}    
 \label{fig:process}  
\end{figure}

Our sky pixel identification system used data from two main sources, the Skyfinder data set \citep{Mihail2016} and GSV \citep{GoogleMaps2017b}. Three computer vision techniques and a number of parameter variations were used to process the data (see Table \ref{tab:techniques}). Two were developed/customised as part of this research. One existing method was implemented and also used as a benchmark. The overall process flow is shown in Figure \ref{fig:process}. Finally, a previously published fourth technique (i.e. Sobel/flood-fill) was also used as a benchmark test for our system.

\begin{table}[!htbp]
\caption{\bf Techniques and variations of each algorithm used in this study for sky pixel detection. \label{tab:techniques}}     
\begin{tabularx}{\textwidth}{ >{\hsize=0.15\hsize}X >{\hsize=0.85\hsize}X }
\hline
\rowcolor{light-gray}\textbf{Sobel} & Implementation of \cite{Wang2015a}'s Sobel operator/hybrid probability model \\
{
\begin{tabularx}{\textwidth}{ X X X X}
\textbf{Variations} & \textbf{Probability threshold} \\ \hline
Sobel\_50 & 0.5 \\    
Sobel\_60 & 0.6 \\     
Sobel\_70 & 0.7 \\    
Sobel\_80 & 0.8 \\     
Sobel\_90 & 0.9 \\  
Sobel\_95 & 0.95 \\
\hline
\end{tabularx}
}
\\
\rowcolor{light-gray}\textbf{Mean shift} & Algorithm developed by the authors based on mean shift segmentation \\
{
\begin{tabularx}{\textwidth}{ X X X X}
\textbf{Variations} & \textbf{Spatial radius (pixels)}&\textbf{Range radius (pixels)}&\textbf{Min. density (pixels)} \\ \hline
Mean\_7\_8\_300 & 7& 8& 300 \\
Mean\_3\_6\_100	& 3& 6& 100 \\
Mean\_5\_7\_210	& 5& 7& 210 \\	 
Mean\_7\_6\_100	& 7& 6& 100 \\
\hline
\end{tabularx}
}
\\
\rowcolor{light-gray}\textbf{K-means} & Algorithm developed by the authors based on K-means clustering and HSL colour filtering \\
{
\begin{tabularx}{\textwidth}{ X X X X X X X X}
\textbf{Variations} & \textbf{Clusters} & \textbf{Skyreq}&\textbf{H$_{high}$}&\textbf{H$_{low}$} & \textbf{L$_{lightness}$} & \textbf{L$_{grey}$}& \textbf{S$_{grey}$} \\ \hline
K-mean\_12 & 12 & 0.7& 0.75& 0.3 & 0.95 & 0.75 & 0.2 \\
K-mean\_6 & 6 & 0.6& 0.75& 0.3 & 0.95 & 0.75 & 0.2 \\
K-mean\_14 & 14 & 0.4& 0.75& 0.3 & 0.95 & 0.65 & 0.2 \\
\hline
\end{tabularx}
}
\\
\rowcolor{light-gray}\textbf{Sobel/flood-fill} & \cite{Middel2018}'s Sobel operator/flood-fill algorithm used as benchmark  \\ \hline
\end{tabularx}
\end{table}

\subsection{Data}\label{sec:data}

\subsubsection{GSV data}\label{sec:gsvdata}
Panoramas for 406 locations in a variety of cities (Adelaide, Brisbane, Paris, Sydney, Tokyo, Perth, and Melbourne) were retrieved using the Google Maps API. Images were retrieved as six 640$\times$640 tiles (one each for up, down, left, right, front, and back directions). The six images were stitched together into a 1280$\times$960 cubic image using Java 8 \citep{Oracle2018} and OpenCV \citep {Bradski2000}. Validation images were created by hand marking sky regions (with blue, RGB 0,0,255) in each image using the GNU Image Manipulation Program \citep{GIMP2019}. Figure \ref{fig:origmarked} shows an example of a GSV panorama image and the corresponding hand-marked validation image. This data set was only used as part of the validation data set.

\subsubsection{Skyfinder data}\label{sec:finderdata}
\begin{figure}
\centering    
\includegraphics[trim={0 0 0 0},clip,scale=0.25]{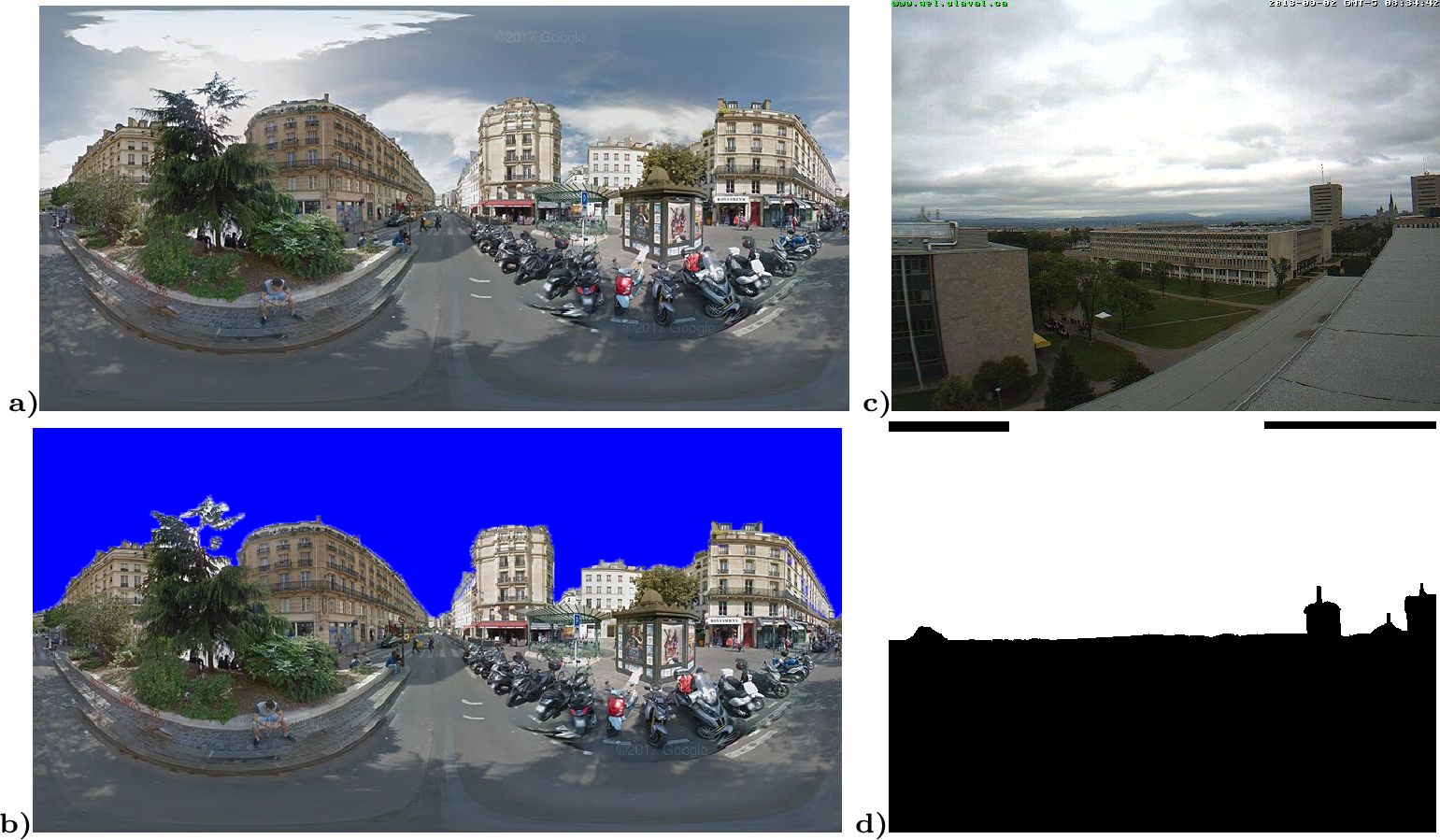} 
\caption{\bf a) Original GSV panorama image and b) hand-marked validation image (sky marked with blue, RGB 0,0,255). c) Original Skyfinder image and d) Skyfinder validation mask (sky marked with white, RGB 255,255,255).}    
 \label{fig:origmarked}  
\end{figure}

This data set was built from 90,000 long-term timelapse images from 53 outdoor webcams over a variety of lighting and weather conditions. Images are of a wide range of sizes and aspect ratios, including 640$\times$489, 857$\times$665, 960$\times$600, 1280$\times$720, and 1280$\times$960. For each location, a binary sky mask was created by \cite{Mihail2016} for validation purposes. An example image and mask (with sky marked with white, RGB 255,255,255) are shown in Figure \ref{fig:origmarked}. All of these images are available from the Skyfinder website \citep{Mihail2015}.

In this study, we selected 38,115 images from 40 locations. Night-time and images with heavy fog were removed as these are conditions unlikely to be encountered in imagery used to calculate SVF. The data set was split into two data sets, 28,586 for neural network training and 9,529 for validation.

\subsection{Techniques and parameters}
\subsubsection{\cite{Wang2015a} Sobel operator/hybrid probability model}\label{sec:prob}
An implementation of the sky detection algorithm presented in \cite{Wang2015a} was implemented using OpenCV and Java 8. This method proceeds by calculating grey-scale gradient images using x- and y-directional Sobel operators to estimate sky colour. An optimised objective function attempts to find the best sky/ground boundary in the gradient image using the covariance matrices of a first calculation of sky and ground regions. Using this best sky boundary, probability models are created from i) the centre and standard deviations of the colours, ii) the gradient values, and iii) the vertical position of each pixel (vertically higher pixels are more likely to be sky). An overall probability model, ranking each pixel's probability (0 to 1) to be sky, is generated from these three probability models. \cite{Wang2015a} reports an error average of 0.051 and standard deviation of 0.058 in their evaluation using human-labelled images.

\begin{figure}
\centering    
\includegraphics[trim={0 0 0 0},clip,scale=0.23]{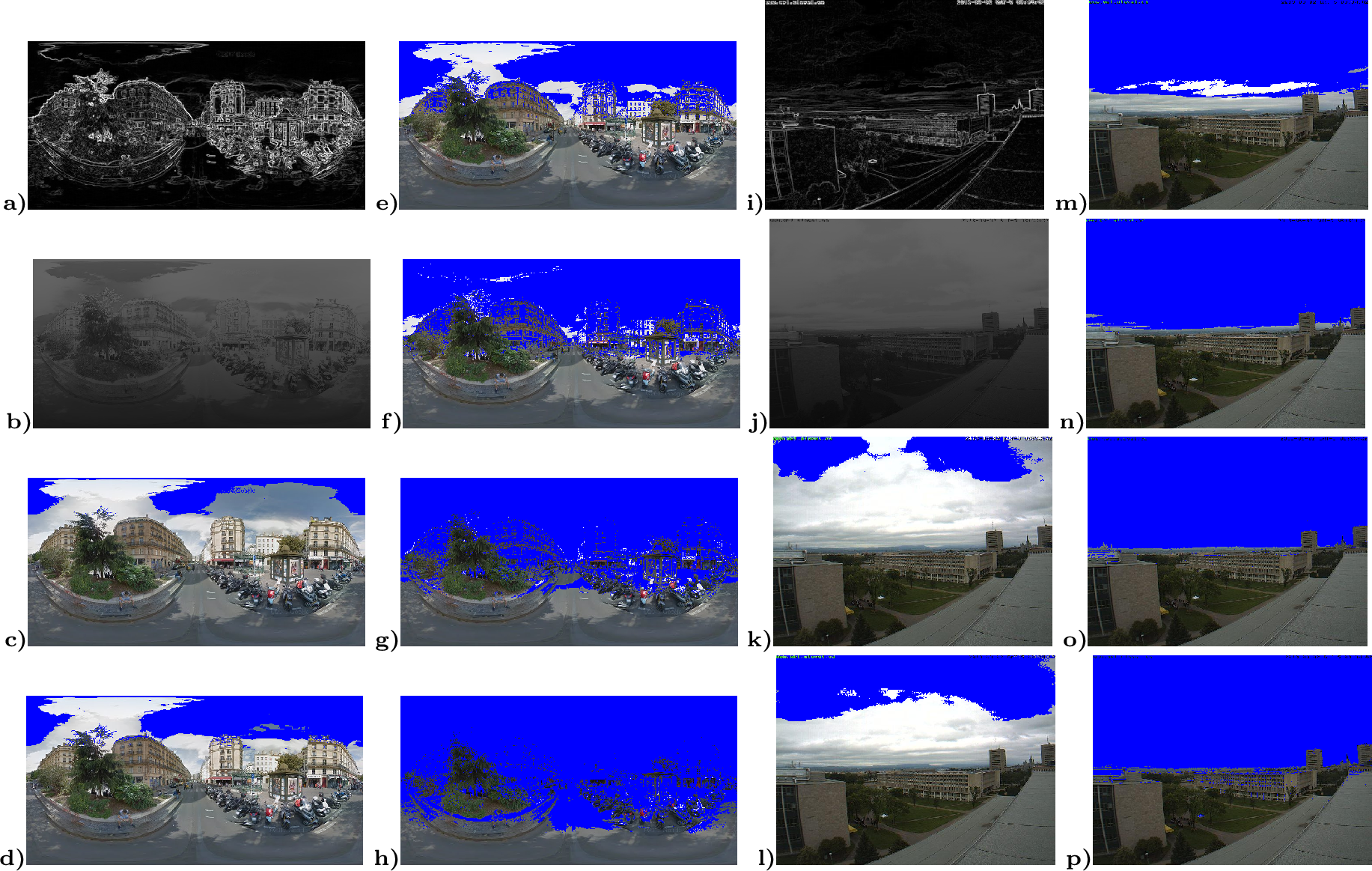}
\caption{\bf Results of Sobel operator/hybrid probability model, showing a) Sobel operator gradient image, b) resulting probability predictions, c) Sobel\_50, d) Sobel\_60, e) Sobel\_70, f) Sobel\_80, g) Sobel\_90, and h) Sobel\_95 for a GSV image. Subfigures i) to p) same as a) to h) but with Skyfinder image.}    
 \label{fig:sobolresults}  
\end{figure}

\cite{Wang2015a} did not recommend a probability threshold, so a number of thresholds were tested (0.5, 0.6, 0.7, 0.8, 0.9, and 0.95) and given the designations of Sobel\_50, Sobel\_60, Sobel\_70, Sobel\_80, Sobel\_90, and Sobel\_95 respectively (Table \ref{tab:techniques}). The algorithm was applied to each image and pixels that exceeded the chosen threshold were marked as sky pixels (using blue, RGB 0,0,255). Results from our implementation are shown in Figure \ref{fig:sobolresults}. Note, two of the best performing variations are also used for a benchmark comparison in Section \ref{sec:sobelwangbenchmark}.

\subsubsection{Mean shift segmentation algorithm}\label{sec:mean}

Mean shift is an algorithm often used for image segmentation \citep{Comaniciu1997,Comaniciu2002}. Image segmentation involves decomposing images into homogeneous contiguous regions of pixels of similar colours or grey levels. Mean shift iteratively picks search windows (spatial and range) of a certain radius in an initial location in an image, computes a mean shift vector, and translates the search window by that amount until convergence \citep{Comaniciu1997}. Segmentation results are highly dependent on input parameters for the algorithm, which include the spatial and colour range radius of the search window and minimum density (the minimum number of pixels to constitute a region). The mean shift used in this project is based on a Java port by \cite{Pangburn2002} of the C++ based EDISON vision toolkit \citep{Christoudias2002}. 

\begin{figure}
\centering 
\includegraphics[trim={0 0 0 0},clip,scale=0.25]{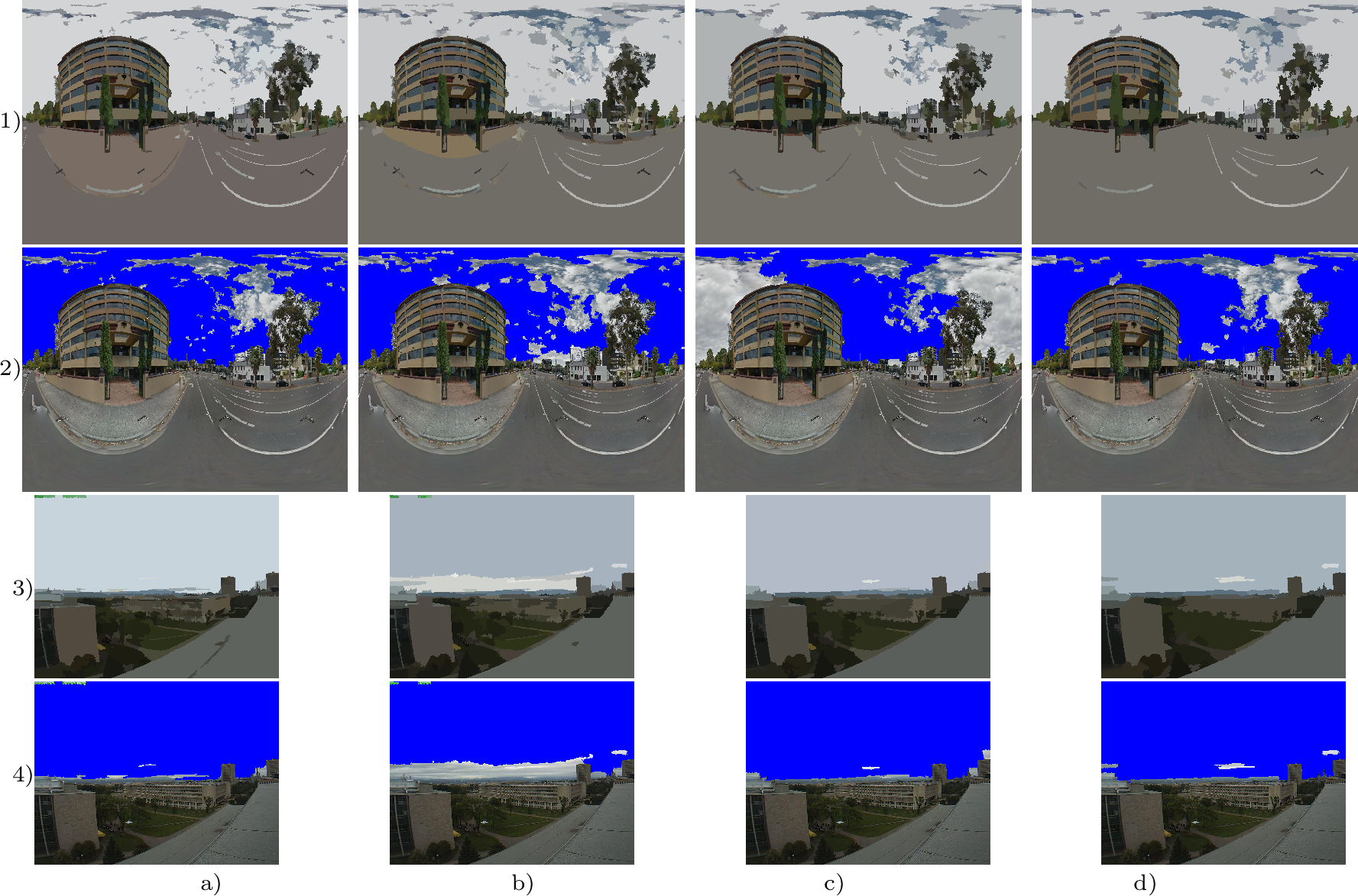} 
\caption{\bf Comparison outputs of intermediate mean shift segmentation algorithm processing steps using varying parameters, showing columns a) Mean\_3\_6\_100, b) Mean\_7\_6\_100, c) Mean\_5\_7\_210, d) Mean\_7\_8\_300 and intermediate mean shifted (GSV row 1, Skyfinder row 3) and the final marked images (GSV row 2, Skyfinder row 4). }
 \label{fig:meanresults}  
\end{figure}

Four different variations of the input parameters were used, determined experimentally through a sensitivity test to work across the widest variety of images. A comparison of a GSV and a Skyfinder location with different mean shift variations and marked results is shown in Figure \ref{fig:meanresults}. The technique designations and parameters are detailed in Table \ref{tab:techniques}. Mean shift is applied to each image with the chosen set of parameters and pixels of the most common colour (in the top half of the segmented image) are marked as sky. 

\begin{figure}
\centering   
\includegraphics[trim={0 0 0 0},clip,scale=0.25]{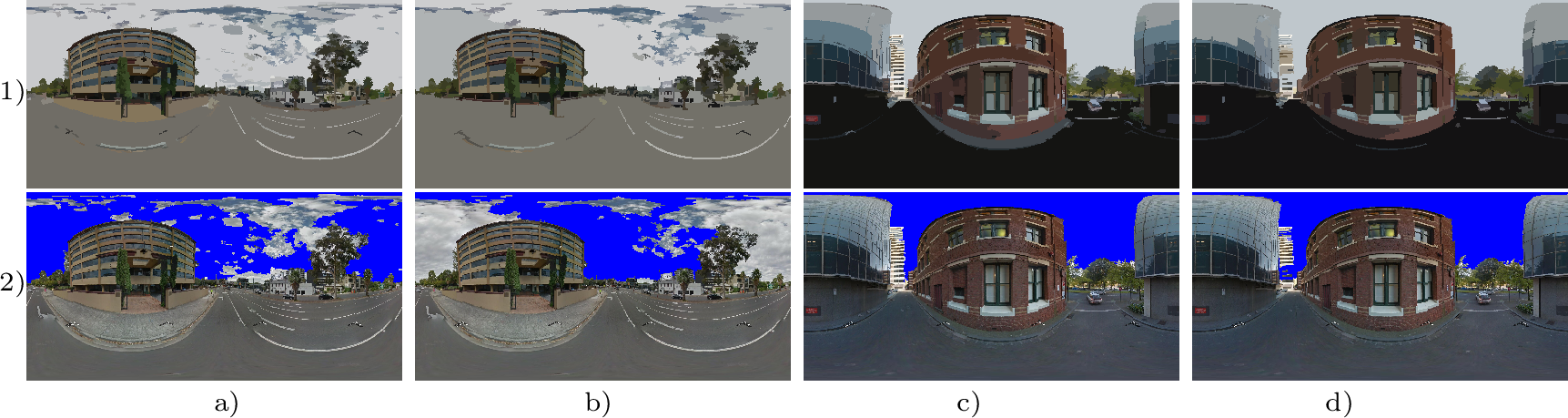} 
\caption{\bf Examples of false negatives and false positives with different mean shift variations at two additional locations of mean shift segmentation parameters. Intermediate mean shifted (row 1) and the final marked images (row 2) of Mean\_7\_6\_100 (columns a and c) and Mean\_5\_7\_210 (columns b and d).}    
 \label{fig:meanerrors}  
\end{figure}

To illustrate the effects of different mean shift variations can have in a variety of images, under varying sky conditions, results from two additional locations are shown in Figure \ref{fig:meanerrors}. To shift the entire sky to a single colour, images with patchy multi-coloured clouds are more accurately segmented when the radius and density parameters are increased (e.g. in column b compared to column a in Figure \ref{fig:meanerrors}). However, at other locations, this can have the effect of creating false positives, for example, where the building in the centre left background is increasingly segmented into the sky in column d in Figure \ref{fig:meanerrors} compared to column c. 

\subsubsection{K-means clustering and HSL color filtering}\label{sec:kmeans}
A third sky segmentation technique was designed using K-means clustering and hue, saturation, and lightness (HSL) colour filtering. K-means clustering iteratively splits an image into $K$ number of clusters, terminating when a specified criteria is met (i.e. maximum iterations and/or desired accuracy). The K-means clustering was performed using the K-means method from the OpenCV library. Three different input parameter settings were used, determined experimentally through a sensitivity test to work on a wide variety of images. The technique designations and parameters are detailed in Table \ref{tab:techniques}. K-mean\_6 was more accurate with cloudy skies. K-mean\_14 handles sky scenes broken up by tree canopies. The last variation, K-mean\_12, generally performed with low accuracy for most images but in a small number of cases (when the sky is mostly obscured by a solid object, building or bridge but not trees) performed better than all the other techniques and variations. 

After K-means clustering, cluster regions were processed based on HSL values. The following conditions (for \textit{H}, hue, \textit{S}, saturation, and \textit{L}, lightness) must be met to add a colour region to a list of possible sky regions: 

$H_{low} < H < H_{high}$

$\cup L > L_{lightness}$

$\cup L > L_{grey} \cap S < S_{grey}$

\begin{figure}
\centering 
\includegraphics[trim={0 0 0 0},clip,scale=0.22]{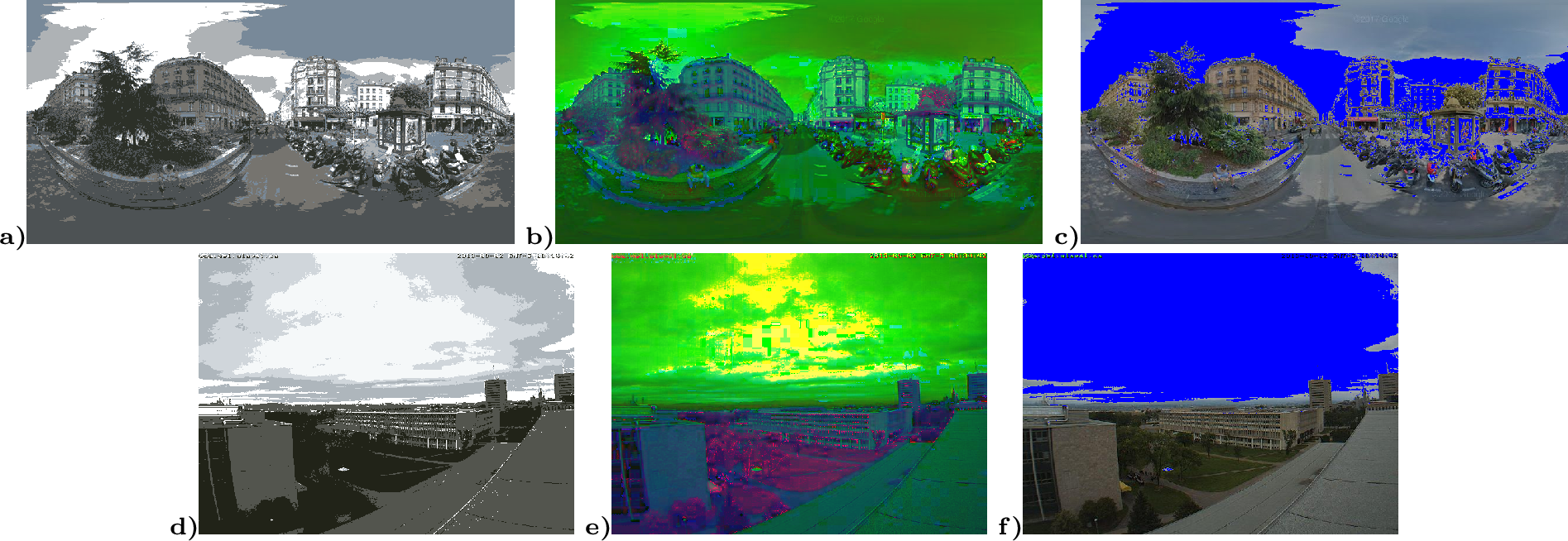}
\caption{\bf Results of K-means clustering and HSL color filtering (K-mean\_6), showing GSV (top row) and Skyfinder (bottom) and K-means clustered image (left), HSL intermediate image (middle), and final marked image (right).}    
 \label{fig:kmeansresults}  
\end{figure}

Of these possible sky clusters, only clusters with a number of pixels greater than the \textit{Skyreq} threshold (percent of all pixels) in the image were finally marked as sky regions. Example results are shown in Figure \ref{fig:kmeansresults}. 

\subsubsection{\cite{Middel2018} Sobel operator/flood-fill algorithm}\label{sec:floodfill}

\begin{figure}
\centering 
\includegraphics[trim={0 0 0 0},clip,scale=0.25]{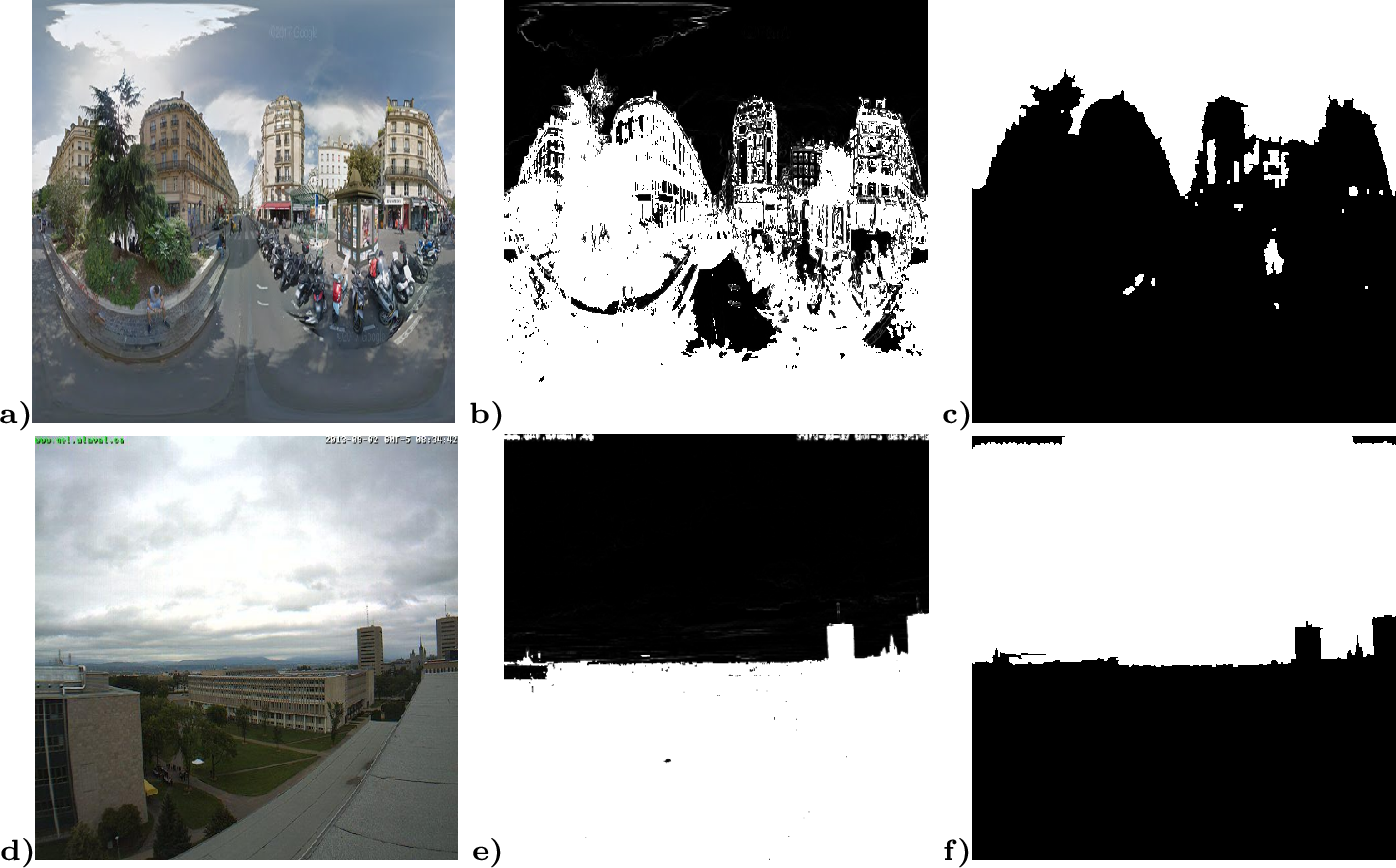} 
\caption{\bf Results of Sobel/flood-fill combination, showing a) original GSV image (rescaled to 512x512), b) intermediate Sobel image, and c) final marked sky image. Subfigures d) to f) same as a) to c) but using Skyfinder imagery.}    
 \label{fig:sobelflood}  
\end{figure}

For benchmark comparisons, we used an algorithm developed by \cite{Middel2018}. The process is based on a Sobel filter \citep{Sobel1968} and flood-fill algorithm \citep{Laungrungthip2008,Middel2017}. The method was designed to calculate SVF from GSV image cubes that were projected into upwards facing fisheye views. Note, this algorithm also rescales the imagery to 512$\times$512. All 38,521 images in the combined training and validation data set were processed with this algorithm (sky pixels marked with white, RGB 255,255,255), compared to validation images, and results saved for a comparison with our results. The results were kept separate from the other 13 techniques and were not included in the NN training process (see Section \ref{sec:nntraining}). Also, in our benchmark comparison, we used this system to process a varied outdoor imagery data set, not the fisheye imagery (cropped below the horizon) the algorithm was originally designed to process. Examples are shown in Figure \ref{fig:sobelflood}.

\subsection{Neural network}\label{sec:nn}

\subsubsection{Inception v3}\label{sec:inception}
\begin{figure}
\centering    
\includegraphics[scale=0.33]{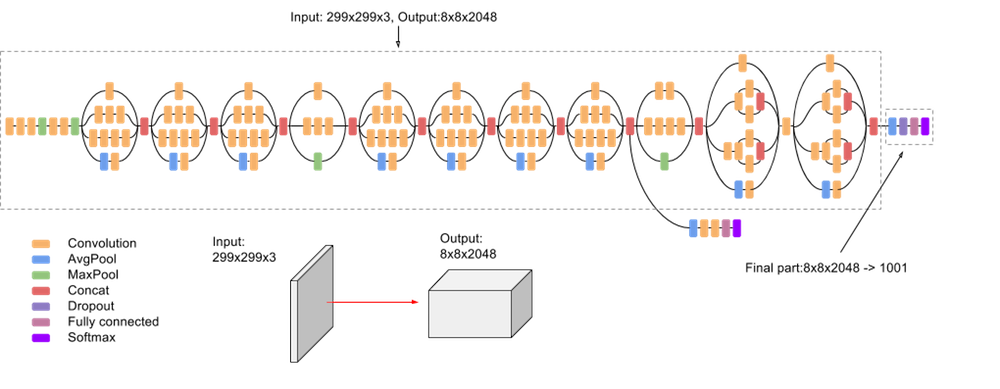}
\caption{\bf Inception v3 architecture \citep{Google2019}.}    
 \label{fig:incetption3}  
\end{figure}

The Microsoft Cognitive Toolkit (CNTK) \citep{Yu2015,Agarwal2016}, with the Inception v3 network \citep{Szegedy2015a}, was used in this project to route images through our adaptive algorithm. This neural network (NN) is a widely used model for image classification across a large variety of fields \citep{Xia2017,Hassannejad2016}. The original Inception model \citep{Szegedy2015} is a deep convolutional neural network (CNN) built using convolutions, average pooling, max pooling, concats, dropouts, and fully connected layers. A reduction of parameters and increased performance over other competing architectures was achieved by the use of Inception modules, combining convolution operations with varying filter sizes. Inception v3 deepens the network (i.e., 42 layers, see Figure \ref{fig:incetption3}) without a large increase in computational requirements by factorising convolution operations with large filter sizes into smaller stacked layers.

\subsubsection{Neural network training}\label{sec:nntraining}    

The Skyfinder data set of 38,115 images was split into two data sets for training (75\%) and validation (25\%). All training and validation images (which consisted of images of a wide variety of sizes and aspect ratios) were preprocessed and rescaled using ImageJ \citep{Rueden2017} to Inception v3's native input size, 299$\times$299, prior to training and validation. 

The network was calibrated using supervised learning, trained with a list of images assigned to categories, in our case, which one of the 13 sky detection technique variations performed most accurately for each image. The training process was run until the model reached peak accuracy (convergence) at recognising these classifications from images. Note, none of the GSV imagery was used in the NN training process.

\subsubsection{Neural network inference}\label{sec:nninference}    
Using the trained model, inferences were performed using the images from the validation data set (the 25\% of images from Section \ref{sec:nntraining}) as well as the 406 GSV panoramas. The techniques and parameters picked by the NN as the most appropriate for that image were used to mark the sky pixels. The marked sky pixels were then compared to the ground truth to assess accuracy.

\section{Results}\label{sec:results}

Three sets of results are reported in this section. The first is a comparison of all the techniques and parameters run individually against the Skyfinder and GSV data sets (a total of 38,521 images). The second presents the results of 9,636 validation images using our adaptive process flow with the techniques and parameters chosen by the trained NN. The third presents a comparison to two benchmark models: a) the \cite{Wang2015a} Sobel operator/hybrid probability model (from which we chose the two best performing variations) and b) the \cite{Middel2018} Sobel operator/flood-fill algorithm.

\subsection{Results from all techniques}\label{sec:resultsall}
\begin{figure}
\centering
\includegraphics[trim={0 0 0 0},clip,scale=0.20]{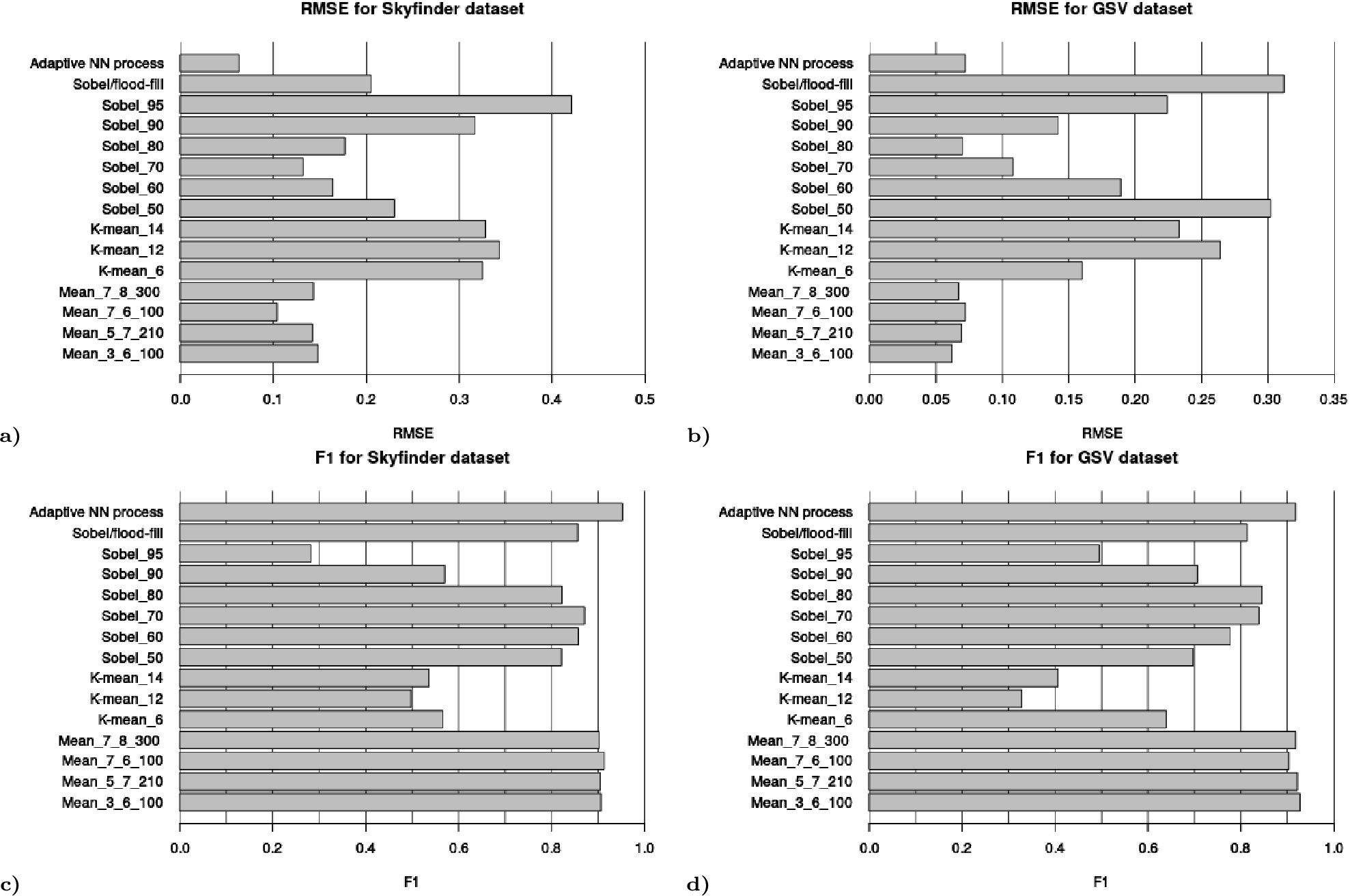} 
\caption{\textbf{
Evaluation of all techniques and parameters showing statistics for RMSE (lower values are more accurate) for a) Skyfinder data set and b) GSV data set. Evaluation of all techniques and parameters showing F1 statistics (higher values are more accurate) for the 9,636 image validation data set, split by c) Skyfinder and d) GSV images.}}
\label{fig:stats}
\end{figure}

\begin{figure}
\centering
\includegraphics[trim={0 0 0 0},clip,scale=0.22]{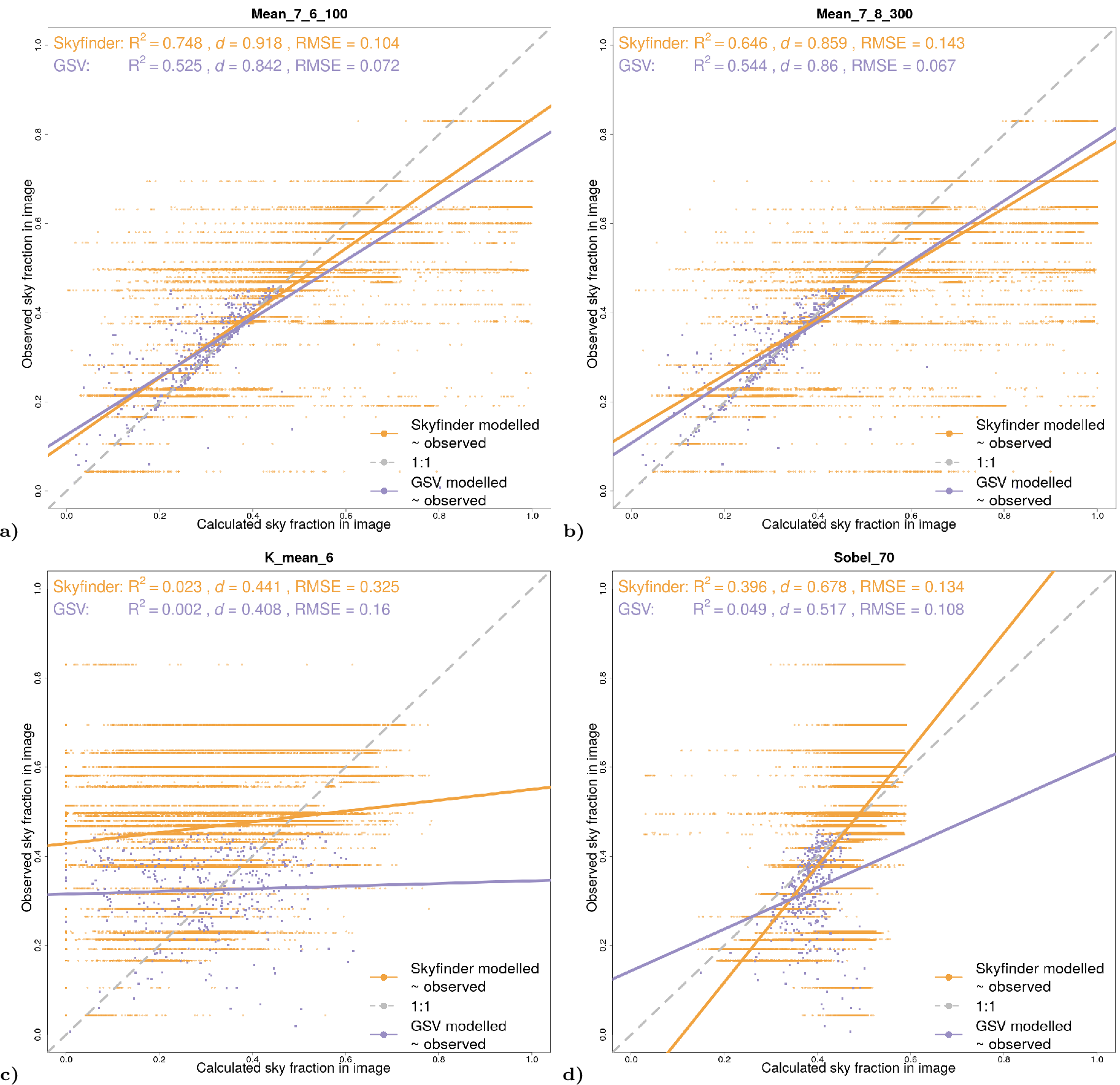} 
\caption{\textbf{Observed vs. calculated sky pixels using the a) Mean\_7\_6\_100, b) Mean\_7\_8\_300, c) K-mean\_6, and d) Sobel\_70 techniques on the 38,521 image combined data set (406 GSV and 38,115 Skyfinder images).} }
\label{fig:errorallcombined}
\end{figure}

All the technique and parameter variations were used to process the two data sets of the 38,115 Skyfinder and 406 GSV images. Figures \ref{fig:stats}a and \ref{fig:stats}b shows a comparison of RMSE statistics for the two data sets. Plots of a number of the better performing techniques are presented in Figure \ref{fig:errorallcombined}. Note, the strong horizontal lines in these figures are due to the nature of the Skyfinder data set that contains large groups of the same scenes (with the same percentage of sky) under different lighting and weather conditions, often resulting in a wide range of predicted results. A complete summary of \cite{Willmott1981}'s index of agreement, \textit{d}, R$^{2}$, and RMSE statistics for evaluations against the two data sets are presented in Supplementary Table \ref{tab:evalall}. F1 statistics for each technique against the 9,636 validation images are presented in Figures \ref{fig:stats}c and \ref{fig:stats}d. A complete summary of precision, recall and F1 statistics are shown in Supplementary Table \ref{tab:precision}.

\subsection{Results from neural network classified techniques}\label{sec:resultsnn}
\begin{figure}
\centering
\includegraphics[trim={0 0 0 0},clip,scale=0.20]{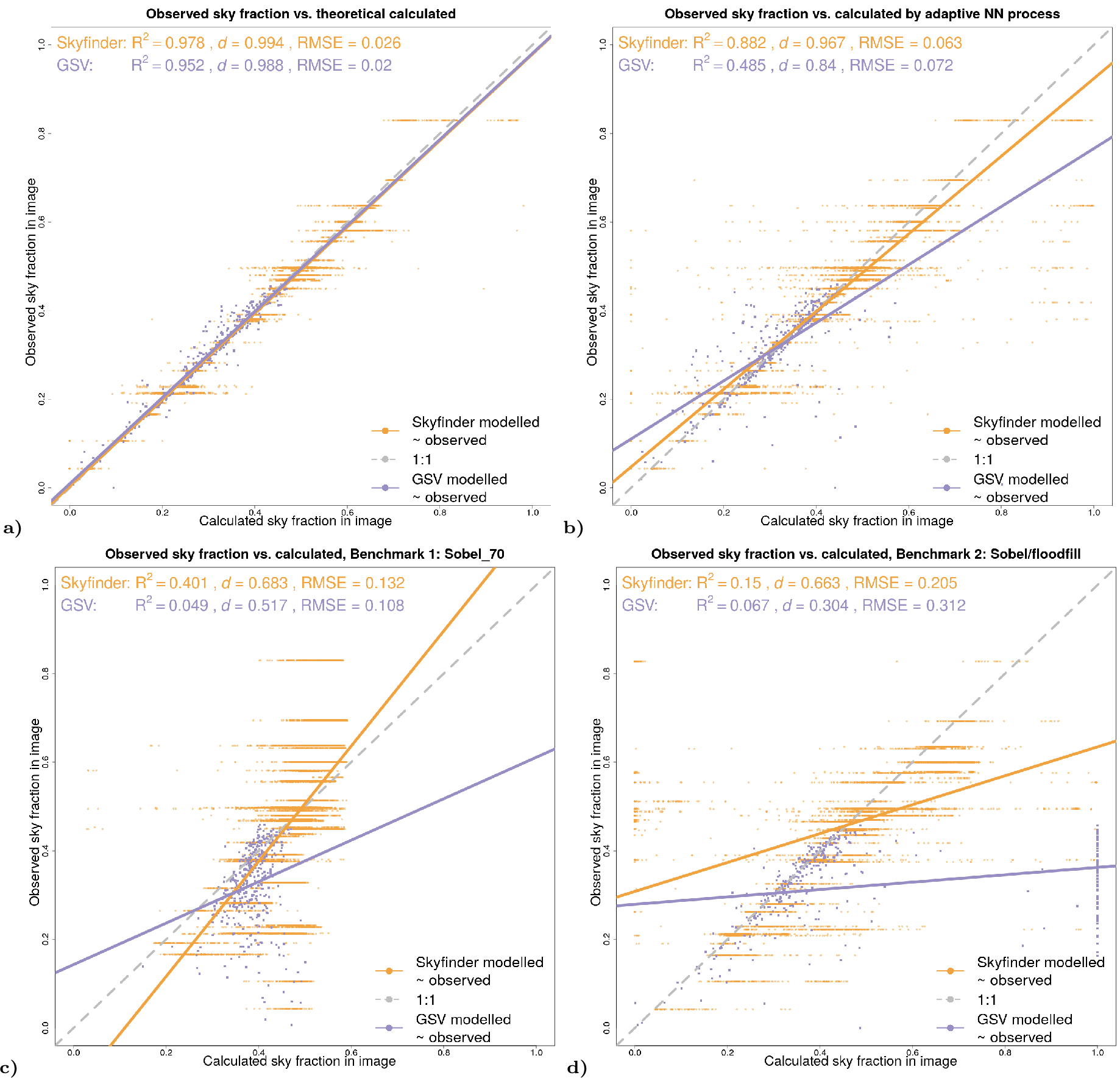} 
\caption{\textbf{a) Theoretical best case results if the NN is 100\% accurate for the 9,636 validation images. Other subfigures show results of b) our adaptive NN process, c) Benchmark 1: Sobel\_70, and d) Benchmark 2: Sobel/flood-fill combination.}}
\label{fig:errorfloodall}
\end{figure}

\begin{figure}
\centering
\includegraphics[trim={0 0 0 0},clip,scale=0.25]{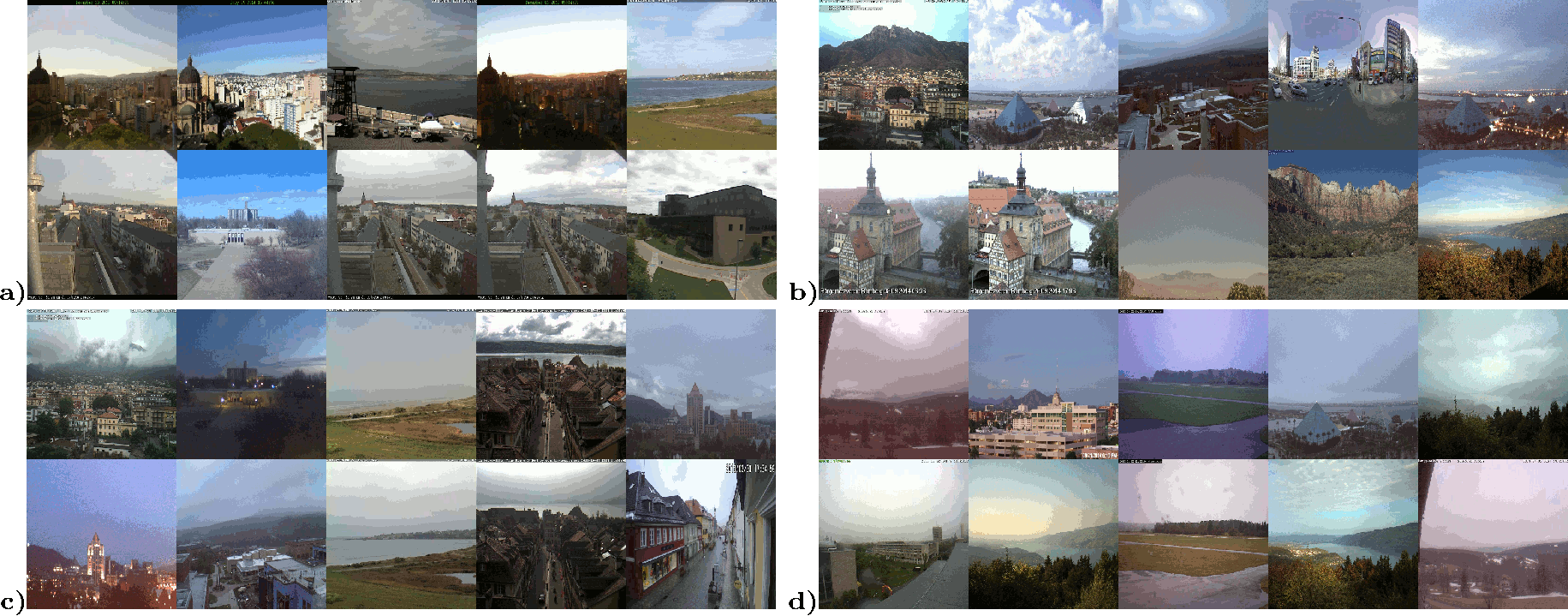} 
\caption{\textbf{Selected imagery used for NN training for classifications 
a) Mean\_7\_8\_300, b) Mean\_7\_6\_100, c) K-mean\_6, and d) Sobel\_70.}}
\label{fig:classImages}
\end{figure}

Figure \ref{fig:errorfloodall}a presents a theoretical best case. If the NN was 100\% accurate in picking the best technique from the 13 possible combinations based on its training, a RMSE of 0.026 and 0.020 (and \textit{d} index of agreement of 0.994 and 0.988) is possible against the validation data set of 9,529 Skyfinder images and 406 GSV images respectively. Samples of the imagery used in training for selected classifications are shown in Figure \ref{fig:classImages}. As can be seen in this figure, there are no strong visual themes in each of the classifications (i.e. all very cloudy, clear blue sky, or multi-coloured sky), however the NN is able to pick up on more subtle features not readily visible to the eye.

The NN was trained for 250 epochs on an Nvidia GeForce GTX 1080 GPU, requiring about 12 hours. The NN reached a peak accuracy rate of 52.6\% in choosing the optimal algorithm from the 13 options. However, as some techniques did only slightly better than others, the error introduced by picking the second or third best algorithm is generally limited. Breaking down the accuracy with this in mind, the NN picked the best method 52.6\% of the time, the second best 18.9\%, and the third best 10.5\% (for a total of the three of 82.0\%)

Figure \ref{fig:errorfloodall}b shows the overall accuracy of the NN chosen pathway process flow against the 9,636 validation images, for the Skyfinder and GSV images (respectively) from the validation data set with a RMSE of 0.063 and 0.072, R$^{2}$ of 0.882 and 0.485, and \textit{d} of 0.967 and 0.840. The accuracy of the NN has impacted the overall accuracy of the system (i.e. not reaching the theoretical accuracy of 0.026 or 0.020 RMSE), but the results on out-of-sample images are still very good. In addition, precision, recall, and F1-scores for the Skyfinder and GSV validation data sets show good results (see Supplementary Table \ref{tab:precision}). Precision is 0.946 and 0.933, recall is 0.965 and 0.918, while the F1-scores are 0.952 and 0.918 (all respectively).

\subsection{Benchmark results}
\subsubsection{Results from the \cite{Wang2015a} Sobel operator/hybrid probability model}\label{sec:sobelwangbenchmark}
To compare the results from our adaptive process (shown in Section \ref{sec:resultsnn}), we picked the Sobel variations that performed best with the two types of imagery in the validation data set. Sobel\_70 (shown in Figure \ref{fig:errorfloodall}c) performed best with the Skyfinder imagery, with a RMSE of 0.132, R$^{2}$ of 0.401, and \textit{d} of 0.683. The best performing variation with the GSV imagery, Sobel\_80 (figure not presented), resulted in a RMSE of 0.07, R$^{2}$ of 0.433 and \textit{d} of 0.655. Precision, recall, and F1-scores statistics (see Supplementary Table \ref{tab:precision}) show that Sobel\_70 has a lower precision than Sobel\_80 with the Skyfinder validation images (at 0.869 vs. 0.920) while Sobel\_70 has better recall scores than Sobel\_80 (0.914 vs. 0.781). Similar patterns are seen with the GSV imagery. Overall, the F1-scores are in the range of 0.82 to 0.87 for both techniques and data sets.

\subsubsection{Results from the \cite{Middel2018} Sobel operator/flood-fill algorithm}\label{sec:resultsflood}
In the evaluation of the Sobel/flood-fill algorithm, results from the Skyfinder and GSV 9,636 validation images are shown in Figure \ref{fig:errorfloodall}d. This algorithm yields a RMSE of 0.205, R$^{2}$ of 0.150 and \textit{d} of 0.663 against the Skyfinder portion of validation data set and a RMSE of 0.312, R$^{2}$ of 0.067, and \textit{d} of 0.304 against with the GSV portion of the validation data set. The results from the evaluation of GSV imagery showed a number of images miss-marked as 100\% sky, inflating the error rate for this data set. A similar problem was seen with the Skyfinder images, several of them were miss-marked as 0\% sky. This is also reflected in the precision, recall, and F1-scores (see Supplementary Table \ref{tab:precision}). Precision scores are 0.840 and 0.761 (for Skyfinder and GSV) are low while the recall scores are much higher (0.900 and 0.948), resulting in overall F1-scores of 0.856 and 0.813. 

\section{Discussion}\label{sec:discussion}

The results from Section \ref{sec:resultsall} show that no single technique and parameter combination performs sky pixel identification with high accuracy across the data sets used by this project. These data sets contain a wide variety of outdoor scenes with various lighting and weather conditions (as can be seen in some of the sample images in Figure \ref{fig:classImages}), challenging many of the techniques. It was expected that algorithms would perform better with GSV imagery, due to their regularity. These images were captured with the same type of equipment, using the same camera angles (horizon at 50\% image height), under clear sky or partly cloudy conditions. The results show that almost all of the variations perform better with the GSV data than the Skyfinder data. Some of the variations even approach the accuracy of our system with the GSV data, for example Sobel\_80. 

However, the Skyfinder data set challenged all of the variations with the Mean and Sobel based methods achieving no better than 0.1 to 0.2 RMSE. However, for some individual images, even the poorest performing techniques (i.e. K-means variations) excelled compared to all of the other techniques. Also, some techniques perform poorly for certain images. In Figure \ref{fig:errorallcombined}d, the results for the Sobel\_70 method show wide variations in $R^{2}$ between the Skyfinder and GSV data sets, while the RMSE values are roughly similar. In the case of Sobel\_70, sky fractions for images with low sky fractions (a small number of images in the data set) are systematically overestimated but this has a significant impact on the GSV $R^{2}$ value. Both of these cases validate the need for an adaptive process that can respond to the specific challenges each image presents to deliver overall better results than any single algorithm. Further, this allows certain techniques to be avoided in the cases that they will perform poorly.

Having a broad range of technique variations was important for the overall accuracy. Experimentation was performed to reduce the number of classifications to possibly increase the accuracy of the NN. However, in removing some of the worst performing methods (many of the K-means variations), the overall accuracy degraded. While some of the variations had very low accuracy overall, they were the best choice for some images. Having those available more than compensated for the NN not always picking the most accurate variations for each image. For example, reducing the choices to the best three performing classes (Mean\_7\_6\_100, K-mean\_6, and Sobel\_70) showed a reduction in RMSE to 0.039 over the best accuracy of the 13 approach combination (RMSE of 0.026 and 0.020 for Skyfinder and GSV respectively). This attempt to increase the accuracy of the NN predictions highlights a limitation of this study. 53\% accuracy in picking between 13 approaches shows that there is room for improvement with this part of the method. With a larger training data set, a lower NN error rate might be achieved, improving performance towards the theoretical RMSE of 0.026 for sky pixel identification. New algorithms and variations of existing algorithms can be added to the system to handle new imagery with greater accuracy, enabled by our flexible framework, as future research. In addition, a NN could be used in a more systematic manner and explore predicting technique parameters directly. 

One difficulty in this study was comparing different sky pixel detection schemes in an objective manner. As noted in the introduction, most studies either do not provide metrics, or provide different types of metrics. Also, with our results evaluating the Skyfinder and GSV data sets showing some large differences in performance, a lack of standardised benchmarks makes comparisons less meaningful. We attempted to overcome these difficulties by implementing some existing methods and including them in our evaluation against common data sets. In addition, the needs of the eventual application should be kept in mind. As precision and recall scores often varied widely for each algorithm, the impact of either higher false positives or false negatives should guide algorithm choice or at least be considered in the results. For example, an algorithm with higher false positives will overestimate sky pixels, leading to a higher SVF estimate and possibly higher maximum temperatures in urban canyon modelling.

\section{Conclusion}\label{sec:conclusion}

In conclusion, we present a system of sky pixel identification that shows high accuracy rates with varied and challenging outdoor imagery. This system sits between algorithms that can be quickly set up and run but are not as accurate with challenging data sets, e.g. \cite{Middel2018}, and more complex systems such as \cite{Gong2018} that require a more extensive deep learning algorithm. Our adaptive system uses the best elements of each in pursuit of the most accurate results possible.

In comparison to published methods, our adaptive process performs well. Our accuracy against the Skyfinder images of RMSE of 0.063 compares well to the RMSE of 0.205 for the \cite{Middel2018} Sobel/flood-fill algorithm and the best performing \cite{Wang2015a} Sobel variations, Sobel\_70 and Sobel\_80, achieving an RMSE of 0.132 and 0.177. Similarly, our adaptive process also performs well on the GSV validation images with a RMSE of 0.072 compared to 0.07 and 0.312 for Sobel\_80 and Sobel/flood-fill respectively. Finally, our adaptive process performs with the best precision, recall, and F1-scores for the Skyfinder data set compared to any of the evaluated techniques.

In the Zenodo data set corresponding to this article (see Section \ref{sec:available}), we provide our trained NN model (and configuration files), which can be used to infer the best algorithm for any type of outdoor imagery, as well as all the training and validation imagery used in this study. This system can then be used out of the box. Using our system, it will be possible to populate databases (such as WUDAPT) of urban morphology information. This data set also provides a standardised data set to reproduce our results and allow benchmark comparisons with other sky pixel detection systems.

\section{Code availability and licensing}\label{sec:available}
Code and data are available from the corresponding author on request. Also, data and code are available at 
https://doi.org/10.5281/zenodo.2562396 and https://bitbucket.org/politemadness/skypixeldetection \citep{Nice2019SkyCode} and are distributed under the Creative Commons Attribution-NonCommercial-ShareAlike 4.0 Generic (CC BY-NC-SA 4.0) license.

\section*{Acknowledgements}
The support of the Commonwealth of Australia through the Cooperative Research Centre program is acknowledged. At Monash University, Kerry Nice was funded by the Cooperative Research Centre for Water Sensitive Cities, an Australian Government initiative. At the University of Melbourne, Kerry Nice was funded by the Transport, Health, and Urban Design (THUD) Hub and a Graham Treloar Fellowship for Early Career Researchers.

\section*{References}\label{sec:ref}

  \bibliographystyle{elsarticle-harv} 
   \bibliography{UrbanClimateICUC10-SVF}

\section{Appendix}\label{sec:app}  
\subsection{Additional data tables}\label{app:tables}  
The following tables provide the underlying data for Figure \ref{fig:stats}.

\begin{table}[!htbp]
\caption{\bf Evaluation of all techniques and parameters for the full Skyfinder and GSV data sets showing statistics for R$^{2}$, RMSE, and \textit{d} index of agreement. \label{tab:evalall}}     
\begin{tabular}{ l l l l l l l }
\multicolumn{1}{c}{\textbf{~}}
& \multicolumn{3}{c}{\textbf{Skyfinder}}
& \multicolumn{3}{c}{\textbf{GSV}}
\\  
 \multicolumn{1}{c|}{\textbf{Designation}}
& \textbf{R$^{2}$} 
& \textbf{RMSE} 
& \multicolumn{1}{c|}{\textbf{\textit{d}}}
& \textbf{R$^{2}$} 
& \textbf{RMSE}  
& \multicolumn{1}{c}{\textbf{\textit{d}}}
\\ \hline

Adaptive NN process&0.882&0.063&0.967&0.485&0.072&0.840 \\
\hline
Sobel/flood-fill&0.124&0.211&0.651&0.067&0.312&0.304 \\
\hline
Sobel\_95       &0    &0.421&0.372&0.053&0.224&0.388 \\
Sobel\_90       &0.041&0.317&0.443&0.214&0.142&0.516 \\
Sobel\_80       &0.255&0.177&0.561&0.433&0.070&0.652 \\
Sobel\_70       &0.396&0.134&0.678&0.049&0.108&0.517 \\
Sobel\_60       &0.288&0.164&0.659&0.005&0.189&0.387 \\
Sobel\_50       &0.104&0.230&0.550&0.026&0.302&0.287 \\
K-mean\_14      &0.029&0.328&0.439&0.023&0.233&0.307 \\
K-mean\_12      &0.028&0.343&0.427&0.017&0.264&0.299 \\
K-mean\_6       &0.023&0.325&0.441&0.002&0.160&0.408 \\
Mean\_7\_8\_300 &0.646&0.143&0.859&0.544&0.067&0.860 \\
Mean\_7\_6\_100	&0.748&0.104&0.918&0.525&0.072&0.842 \\
Mean\_5\_7\_210	&0.659&0.142&0.862&0.526&0.069&0.852 \\
Mean\_3\_6\_100	&0.658&0.148&0.854&0.603&0.062&0.866 \\
\hline
\end{tabular}
\end{table}

\begin{table}[!htbp]
\caption{\bf Evaluation of all techniques and parameters for the 9,636 image validation data set, split by Skyfinder and GSV images, showing statistics for precision, recall, and F1 scores. \label{tab:precision}}     
\begin{tabular}{ l l l l l l l }
\multicolumn{1}{c}{\textbf{~}}
& \multicolumn{3}{c}{\textbf{Skyfinder}}
& \multicolumn{3}{c}{\textbf{GSV}}
\\ 
 \multicolumn{1}{c|}{\textbf{Designation}}
& \textbf{Precision} 
& \textbf{Recall} 
& \multicolumn{1}{c|}{\textbf{F1}}
& \textbf{Precision} 
& \textbf{Recall}  
& \multicolumn{1}{c}{\textbf{F1}}
\\ \hline
Adaptive NN process&0.946&0.965&0.952&0.933&0.918&0.918 \\
\hline
Sobel/flood-fill&0.840&0.900&0.856&0.761&0.948&0.813 \\
\hline
Sobel\_95       &0.921&0.181&0.282&0.931&0.349&0.495\\
Sobel\_90       &0.936&0.436&0.570&0.916&0.586&0.706\\
Sobel\_80       &0.920&0.781&0.822&0.866&0.839&0.845\\
Sobel\_70       &0.869&0.914&0.871&0.788&0.925&0.839\\
Sobel\_60       &0.807&0.967&0.857&0.675&0.959&0.777\\
Sobel\_50       &0.742&0.987&0.821&0.565&0.973&0.697\\
K-mean\_14      &0.708&0.484&0.536&0.640&0.349&0.406\\
K-mean\_12      &0.677&0.441&0.497&0.547&0.272&0.328\\
K-mean\_6       &0.658&0.538&0.566&0.694&0.663&0.639\\
Mean\_7\_8\_300 &0.879&0.944&0.902&0.941&0.909&0.918\\
Mean\_7\_6\_100	&0.912&0.930&0.913&0.951&0.876&0.903\\
Mean\_5\_7\_210	&0.879&0.949&0.905&0.946&0.911&0.921\\
Mean\_3\_6\_100	&0.876&0.956&0.906&0.952&0.915&0.927\\
\hline
\end{tabular}
\end{table}

\end{document}